\documentclass[runningheads]{llncs}
\usepackage{graphicx}
\usepackage{orcidlink}
\usepackage[sort&compress,numbers]{natbib}
\usepackage{multirow}
\usepackage{graphicx}
\usepackage{booktabs}

\begin{document}
\title{Beqi: Revitalize the Senegalese Wolof Language with a Robust Spelling Corrector\thanks{Supported by the Google PhD Fellowship program}}
\titlerunning{Wolof Spelling Corrector}
%

\author{Derguene Mbaye\inst{1,2}\orcidlink{0000-0002-7490-2731} \and
Moussa Diallo\inst{2} }
%
\authorrunning{D. Mbaye et al.}
%
\institute{Baamtu, Dakar, S\'{e}n\'{e}gal\\
\email{derguene.mbaye@baamtu.com}\\
\url{https://www.baamtu.com/} \and
Universit\'{e} Cheikh Anta Diop, Dakar, S\'{e}n\'{e}gal\\
\email{moussa.diallo@esp.sn}}
\maketitle              
\begin{abstract}
The progress of Natural Language Processing (NLP), although fast in recent years, is not at the same pace for all languages. African languages in particular are still behind and lack automatic processing tools. Some of these tools are very important for the development of these languages but also have an important role in many NLP applications. This is particularly the case for automatic spell checkers. Several approaches have been studied to address this task and the one modeling spelling correction as a translation task from misspelled (noisy) text to well-spelled (correct) text shows promising results. However, this approach requires a parallel corpus of noisy data on the one hand and correct data on the other hand, whereas Wolof is a low-resource language and does not have such a corpus.
In this paper, we present a way to address the constraint related to the lack of data by generating synthetic data and we present sequence-to-sequence models using Deep Learning for spelling correction in Wolof. We evaluated these models in three different scenarios depending on the subwording method applied to the data and showed that the latter had a significant impact on the performance of the models, which opens the way for future research in Wolof spelling correction.

\end{abstract}
\section{Introduction}
Spelling mistakes are common in language usage and can be due to a lack of language skills or carelessness. They can become an important element to take into account when writing emails, speeches or when searching on the internet. This is the reason why automatic correctors can be found in various NLP applications such as Summarization \cite{ELKASSAS2021113679}, Machine Translation \cite{mt} and Search Engines \cite{GSudeepthi2012ASO}. Regarding Wolof specifically, it is a language that is more spoken than written, like most African languages. 
The Wolof alphabet has been defined by presidential decree since 1971\footnote{\it{Decree No. 71-566 of May 21st, 1971 concerning the transcription of national languages. Republic of Senegal, 1971}} as well as spelling and word separation in 2005\footnote{\it{Decree no. 2005-992 of October 21st, 2005 concerning the spelling and separation of words in Wolof (currently effective)}} but its adoption remains weak.
Although it is the predominant language spoken in Senegal (statistically), Wolof is not taught in school as it has been supplanted by French, the official language since colonization. All these aspects contribute to the fact that the majority of the population has a weak grasp of the writing of this language and it is common to note spelling mistakes on social networks, advertising posters and even in television programs. Nevertheless, in recent years there has been a significant resurgence of interest in the language and several initiatives to revitalize it have been launched. A group of linguists called WAX ("Wolof Ak Xamle" meaning Wolof and knowledge sharing) has been created and is working on the popularization of Wolof\footnote{\href{https://afrique.le360.ma/senegal/societe/2020/10/03/32064-senegal-le-travail-de-titan-des-academiciens-de-la-langue-wolof-32064/}{Senegal: The Titan work of Wolof language academics, by \it{le360 Afrique} (French)}} by content creation and the launch of an e-learning platform\footnote{\url{https://jangwolof.com/}}, among other things. All these initiatives contributed greatly to the acceleration of the adoption of the written form of this language.

\begin{figure}[htbp]
\centerline{\includegraphics[width=1\textwidth, keepaspectratio]{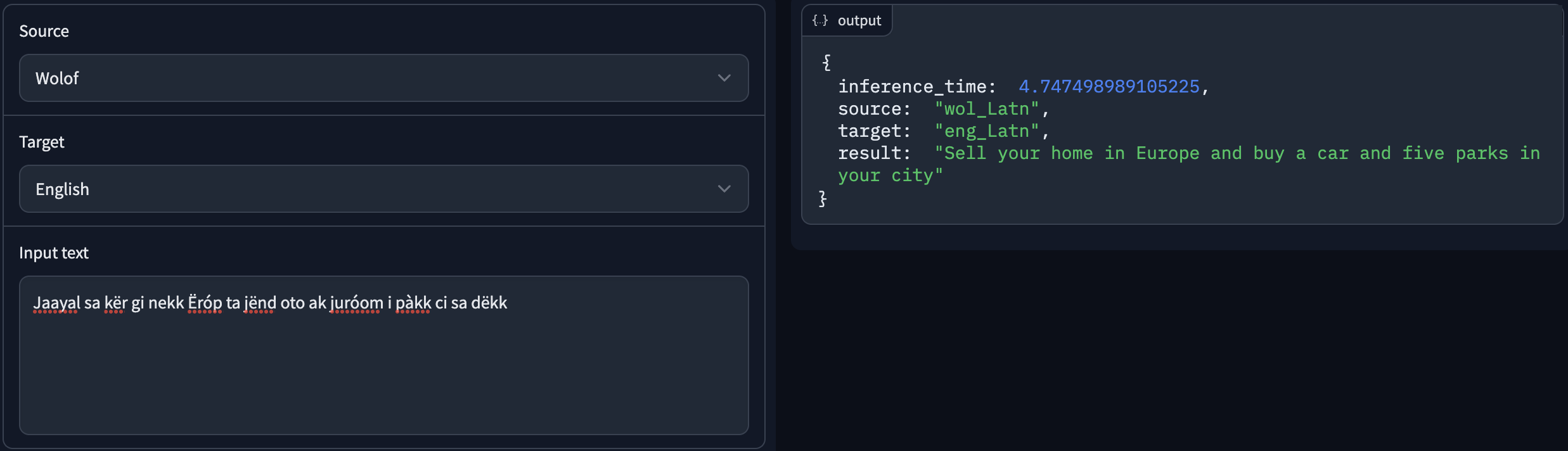}}
\caption{A correctly done translation when the sentence in Wolof is written with the official form. The correct translation of $\texttt{pàkk}$ would be $\texttt{plot of land}$ but the overall meaning is maintained.}
\label{nllb-wolof-good}
\end{figure}

\begin{figure}[htbp]
\centerline{\includegraphics[width=1\textwidth, keepaspectratio]{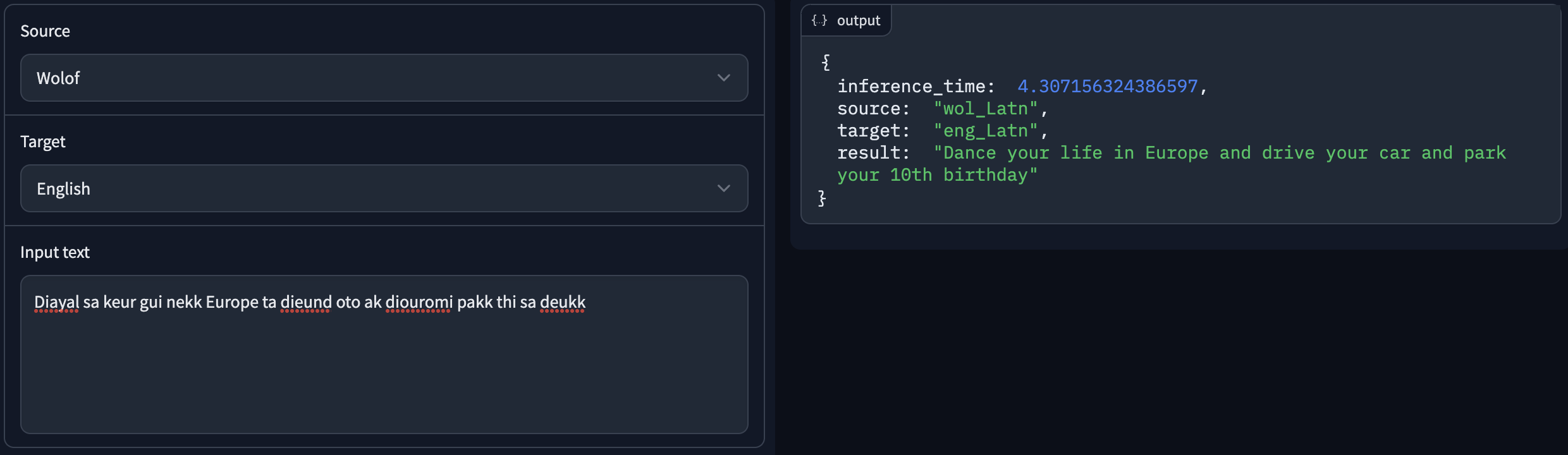}}
\caption{A totally wrong translation when the same sentence is written with the conventional form.}
\label{nllb-wolof-bad}
\end{figure}

However, The incorrect writing has become so democratized that we can consider them as an orthographic system to which we will refer in this article by the term "conventional form". The one based on the official spelling will be called "Official Form". The existence of these two forms of writing creates a gap that can greatly hinder the performance of NLP applications designed for Wolof. In fact, the data sets collected to date in Wolof \cite{TIEDEMANN12.463, strassel-tracey-2016-lorelei, adelani-etal-2021-masakhaner, goyal-etal-2022-flores} are based on the official alphabet and the spelling used is different from the conventional form that is commonly used by the population. NLP applications designed from these datasets will therefore have a lot of trouble working once in production due to this gap. Fig. \ref{nllb-wolof-good} and Fig. \ref{nllb-wolof-bad} illustrate this problem with the translation system designed in \cite{nllb} by Meta researchers\footnote{The translations were performed with the NLLB model distilled to 600M parameters}.

It is thus crucial to have a spell checker in Wolof in order to bridge the gap between the conventional form and the official one. Wolof being a low-resource language, it makes this task even more challenging.

It is in this context that we introduce \textsc{Beqi}\footnote{A Wolof word meaning the action of correcting}: the first Deep Learning-based Wolof spelling corrector for end-to-end learning. We structured the paper as follows:
\begin{itemize}
  \item We begin by presenting the work done in automatic spell correction in Wolof and other low-resource languages in Section \ref{rel-work}.
  \item Data collection and synthetic data generation are discussed in Section \ref{data}.
  \item In Section \ref{exp}, we present the model used and the experiments.
  \item Section \ref{results} presents the results and perspectives.
  \item The conclusion is presented in section \ref{concl}.
\end{itemize}

\section{Related Work}\label{rel-work}
Several approaches have been studied to address the problem of automatic spelling correction in general. The study in \cite{survey} divides these approaches into three groups: 
\begin{enumerate}
  \item One that is based on expert rules ;
  \item One that adds a context model to rearrange candidate corrections ;
  \item One that learns error patterns from a set of training data.
\end{enumerate}

A portable spellchecker for the Amharic language, spoken in Ethiopia, was developed in \cite{gezmu-etal-2018-portable}. The system uses a corpus-driven approach that uses a noisy channel to derive linguistic knowledge to correct spelling errors. Grammatical error correction in low-resource scenarios was studied in \cite{naplava-straka-2019-grammatical} with a focus on the Czech language. The researchers modeled the correction task as a machine translation task with a Transformer-based model \cite{NIPS2017_3f5ee243}. Indian languages have also been studied in the spelling correction task in particular in \cite{etoori-etal-2018-automatic} which uses a Deep Learning based approach and targets Hindi and Telugu languages. Their approach also leverages the machine translation framework and uses a sequential encoder-decoder model based on the Long short-term memory (LSTM) architecture \cite{lstm}. 

Although significant work has been done in spelling correction in low-resource languages, little work has been done in this area for Wolof specifically. Several dictionaries were developed in the context of the Dictionnaires Langue Africaine-Français (DiLAF) project which covered five other African languages in addition to Wolof \cite{Cheikh2015}. The implementation of a spellchecker for Wolof was studied in \cite{lo:hal-02054917} with an approach based on a French-Wolof dictionary studied in \cite{Khoule2016iBaatukaay} as a lexicon and a morphological analyzer of the Wolof language explored in \cite{dione-2012-morphological}. But the work did not go as far as the actual implementation of a functional corrector and was limited to the state of the art of methods based on the first two previously mentioned approaches i.e. those based on expert rules and those using a context model based on n-gram language models. In addtion, at the time of writing this article, all dictionaries developped in \cite{Cheikh2015} are available online\footnote{\href{http://pagesperso.ls2n.fr/~enguehard-c/DiLAF/index.php}{Website of the DILAF project}} except Wolof, which prevents us from exploring a dictionary-based approach. These are also difficult to maintain (the number of rules can quickly increase and their update is tedious), are limited by the size of the dictionary and do not take into account the context. The latter can be included thanks to a context model which is generally an n-gram language model \cite{langmodel} that defines the probability according to the history of the words. This language model thus only takes into account the previous words in addition to the current word, which limits the context considered. Although additional classifiers can be used to bridge this gap in context \cite{survey}, the use of neural networks allows the inclusion of a broader context on both sides of a word.

Deep Learning is thus a promising approach that has been studied for the spelling correction task and for different languages. But to the best of our knowledge, this is the first attempt applied to the Wolof language.

\section{Data Collection}\label{data}

\begin{figure}[htbp]
\centerline{\includegraphics[width=1\textwidth, keepaspectratio]{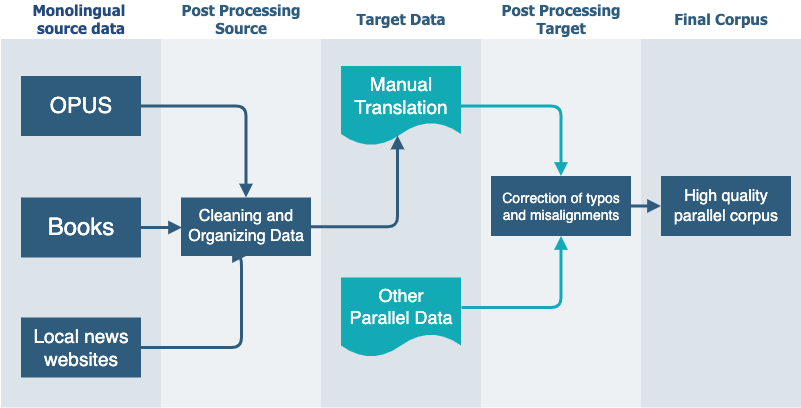}}
\caption{Data collection pipeline}
\label{data-collection}
\end{figure}

\begin{figure}[htbp]
\centerline{\includegraphics[width=1\textwidth, keepaspectratio]{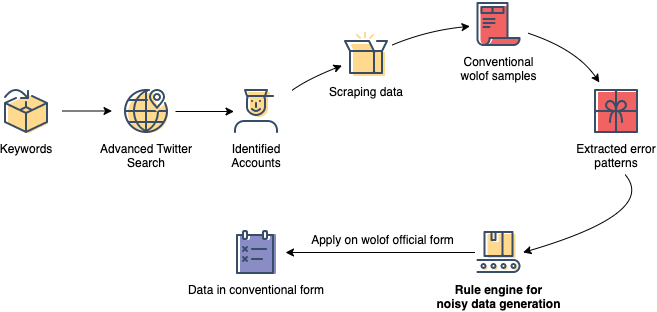}}
\caption{The collection steps that led to the generation of the parallel corpus with noisy data as source and correctly written data as target.}
\label{collection}
\end{figure}

We have collected an in-house dataset of 154,000 correctly written sentences in Wolof which is an extension of the dataset presented in \cite{mbaye2023lowresourced}. These sentences were obtained by first collecting monolingual French data from various sources: Coran, Bible, books and news sites as illustrated in Fig.\ref{data-collection}. Since Senegal is a French-speaking country, it is easier to find linguists who master both languages (Wolof and French) in order to make the best possible translations. We thus collaborated with a team of linguists from the Linguistic Department of the Cheikh Anta Diop University of Dakar to manually translate the collected French corpus into Wolof. The Wolof corpus thus collected and written in the official form constitutes the "target language" that we wish to have as output. To obtain the data of the "source language" written in conventional Wolof, we scraped data on Twitter from accounts that generally publish in Wolof in order to detect recurrent spelling error patterns. Indeed, Twitter is a micro-blogging platform where people write casually about various topics. This makes it an ideal candidate for collecting data that may contain spelling errors and the platform is much in demand for data collection for NLP tasks such as Sentiment Analysis \cite{DRUS2019707}. Author accounts of Wolof publications were identified using Twitter's advanced search functionality by searching for conventional Wolof keywords that appear in tweets and picking up the corresponding authors. From there, we scrape a sample of tweets and identify patterns of errors that we will subsequently reproduce on our corpus written in official form. The overall collection process is illustrated in Fig. \ref{collection}.

\begin{figure}[htbp]
\centerline{\includegraphics[width=1\textwidth, keepaspectratio]{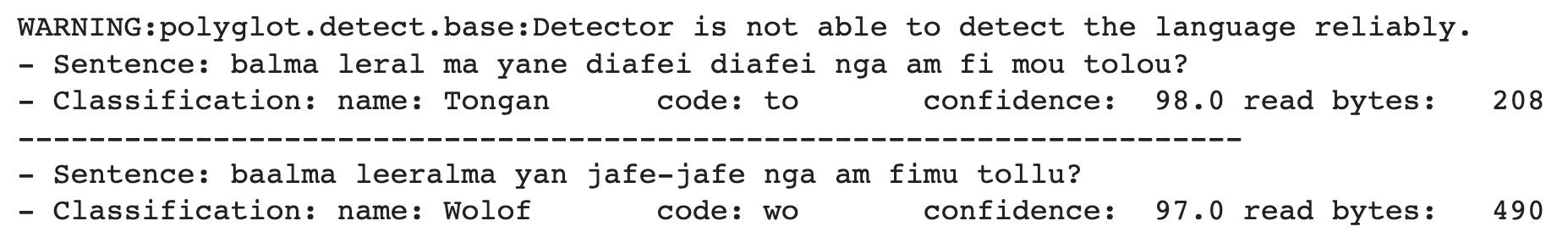}}
\caption{The model failing to recognize the text language when written in the conventional form (above the dash line) and succeeding when written in the official form.}
\label{polyglot-bad}
\end{figure}

However, since people are generally bilingual, many publications are also written in French which includes artifacts in the collected data. To filter them, we first used the language identification model \cite{hughes-etal-2006-reconsidering} included in the polyglot library\footnote{\url{http://www.polyglot-nlp.com/}} in its $\it{16.7.4}$ version to detect the languages of the tweets in order to remove those in French. However, we encountered the problem illustrated in Fig. \ref{nllb-wolof-bad} where the model struggles to detect the language when the text is written in conventional form as illustrated in Fig. \ref{polyglot-bad}. We thus had to do the filtering manually.

Once samples of texts in conventional form were collected, we designed a rule engine based on regular expressions\footnote{Patterns used to match character combinations in strings} where each identified error pattern is transcribed into a defined rule to be applied on the corpus. For example, in the conventional sentence $\texttt{"Diappal bal bi"}$ meaning $\texttt{"Catch the ball"}$, we derive the rule that the "J" followed by a vowel (except "i") is often wrongly replaced by the string "Di". Thus the correct writing of the previous sentence is $\texttt{"Jàppal bal bi"}$. We reproduce this error in our corpus by replacing all the times where the letter "J" is followed by a vowel that is not "i", by the string "Di". The Table \ref{rule-engine-tab} presents the rules used in the engine when pre-processing the data. Two other rules were applied in a post-processing phase: one to remove spaces between a word and a vowel (used in formal Wolof to express a plurality for example) and another one to replace occurences of 'g' followed by vowels by 'gu'. All this process allowed us to collect as much synthetic data as formal data i.e. 154,000 parallel sentences of noisy text on one side and well written text on the other, that will be used to train the final spelling correction model. Some examples of the resulting output of this transformation are shown in Table \ref{transf-examples}.

\begin{table}[htbp]
\setlength{\tabcolsep}{4pt}
\renewcommand{\arraystretch}{1.2}
\caption{Patterns used in regular expressions to map the correct writing to manually identified errors on the collected data ("f/b" means "followed by").}
\begin{center}
\begin{tabular}{c c c}
\hline\rule{0pt}{12pt}
\textbf{Patterns} & \textbf{Replacement} & \textbf{Description} \\[2pt]
\hline\rule{0pt}{12pt}
ñ+ & gn & Replace occurences of 'ñ' by 'gn' \\
\hline\rule{0pt}{12pt}
$\eta$+ & ng & Replace occurences of '$\eta$' by 'ng' \\
\hline\rule{0pt}{12pt}
ë+ & eu & Replace occurences of 'ë' by 'eu' \\
\hline\rule{0pt}{12pt}
u+ & ou & Replace occurences of 'u' by 'ou' \\
\hline\rule{0pt}{12pt}
u([blt]+) & ou\textbackslash1 & Replace occurences of 'ub/l/t' by 'oub/l/t' \\
\hline\rule{0pt}{12pt}
q & kh & Replace every 'q' character by 'kh' \\
\hline\rule{0pt}{12pt}
x & kh & Replace every 'x' character by 'kh' \\
\hline\rule{0pt}{12pt}
u\textbackslash b & ou & Replace words ended with 'u' by 'ou' \\
\hline\rule{0pt}{12pt}
c([aeiouy]\{1,\}) & th\textbackslash1 & Replace occurences of 'c' f/b vowels by 'th' \\
\hline\rule{0pt}{12pt}
c\{2\}\textbackslash b & thie & Replace 'cc' at the end of a word by 'thie' \\
\hline\rule{0pt}{12pt}
[Jj]([eao]{1,2}) & di\textbackslash1 & Replace 'j' f/b a vowel (except i and u) by 'di' \\
\hline\rule{0pt}{12pt}
[Jj]([i]+) & dj\textbackslash1 & Replace 'j' f/b occurences of 'i' by 'dj' \\
\hline\rule{0pt}{12pt}
[Jj]([u]+) & dio\textbackslash1 & Replace 'j' f/b occurences of 'u' by 'dio' \\
\hline\rule{0pt}{12pt}
th([aeouy]+) & thi\textbackslash1 & Replace occurences of 'th' f/b vowels (except i) by 'th' \\[2pt]
\hline\rule{0pt}{12pt}

\end{tabular}
\label{rule-engine-tab}
\end{center}
\end{table}

\begin{table}[htbp]
\setlength{\tabcolsep}{4pt}
\renewcommand{\arraystretch}{1.2}

\caption{Examples of sentences in the official form converted to conventional form by the rule engine.}
\begin{center}
\begin{tabular}{c c}
\hline\rule{0pt}{12pt}
\textbf{Official form} & \textbf{Conventional form} \\[2pt]
\hline\rule{0pt}{12pt}
Nàngul kula raw, kula ëppalé & Nangoul koula raw, koula euppale\\ 
\hline\rule{0pt}{12pt}
Kula gën a taaru ak kula mag & Koula gueuna taarou ak koula mag  \\
\hline\rule{0pt}{12pt}
Yii yëpp dula wàññi dara & Yii yeupp doula wagni dara  \\
\hline\rule{0pt}{12pt}
Wànté bul nangu mukk kula gën & Wante boul nangou moukk koula gueun  \\
\hline\rule{0pt}{12pt}
Lilakoy may, mooy nga sàmm sa ngor & Lilakoy may, mooy ngua samm sa ngor  \\

\hline\rule{0pt}{12pt}

\end{tabular}
\label{transf-examples}
\end{center}
\end{table}

\section{Experiments}\label{exp}

\begin{figure}[htbp]
\centerline{\includegraphics[width=1\textwidth, keepaspectratio]{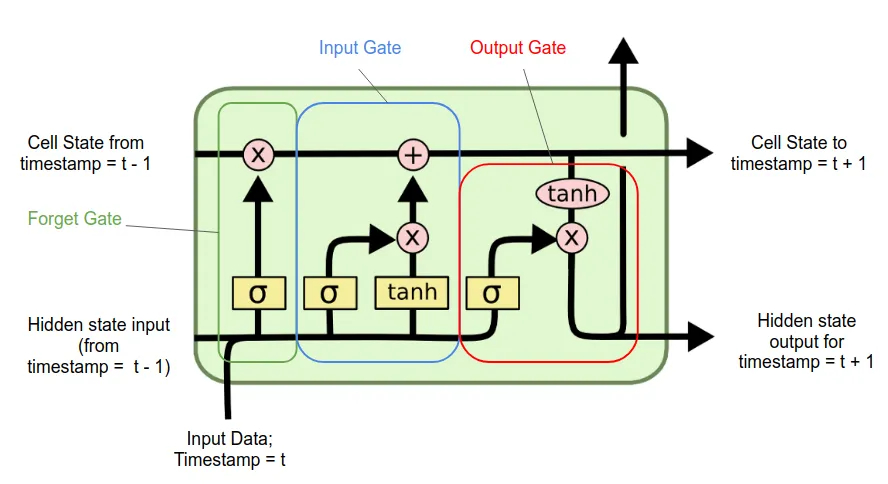}}
\caption{Illustration of an LSTM cell}
\label{lstm}
\end{figure}

When designing the spelling correction system we considered two architectures commonly used in sequence-to-sequence mapping tasks: the LSTM \cite{lstm} and the Transformer\cite{NIPS2017_3f5ee243}. LSTMs are a particular type of Recurrent Neural Networks (RNNs) \cite{rnns}, consisting of several gates that allow them to manipulate the information flow. This manipulation is performed by forgetting or selectively memorizing the information of the previous temporal sequence in a dynamic memory as shown in  Fig.\ref{lstm}\footnote{\href{https://medium.com/analytics-vidhya/lstms-explained-a-complete-technically-accurate-conceptual-guide-with-keras-2a650327e8f2}{LSTMs Explained: A Complete, Technically Accurate, Conceptual Guide with Keras}}. 

\begin{figure}[htbp]
\centerline{\includegraphics[width=0.6\textwidth, keepaspectratio]{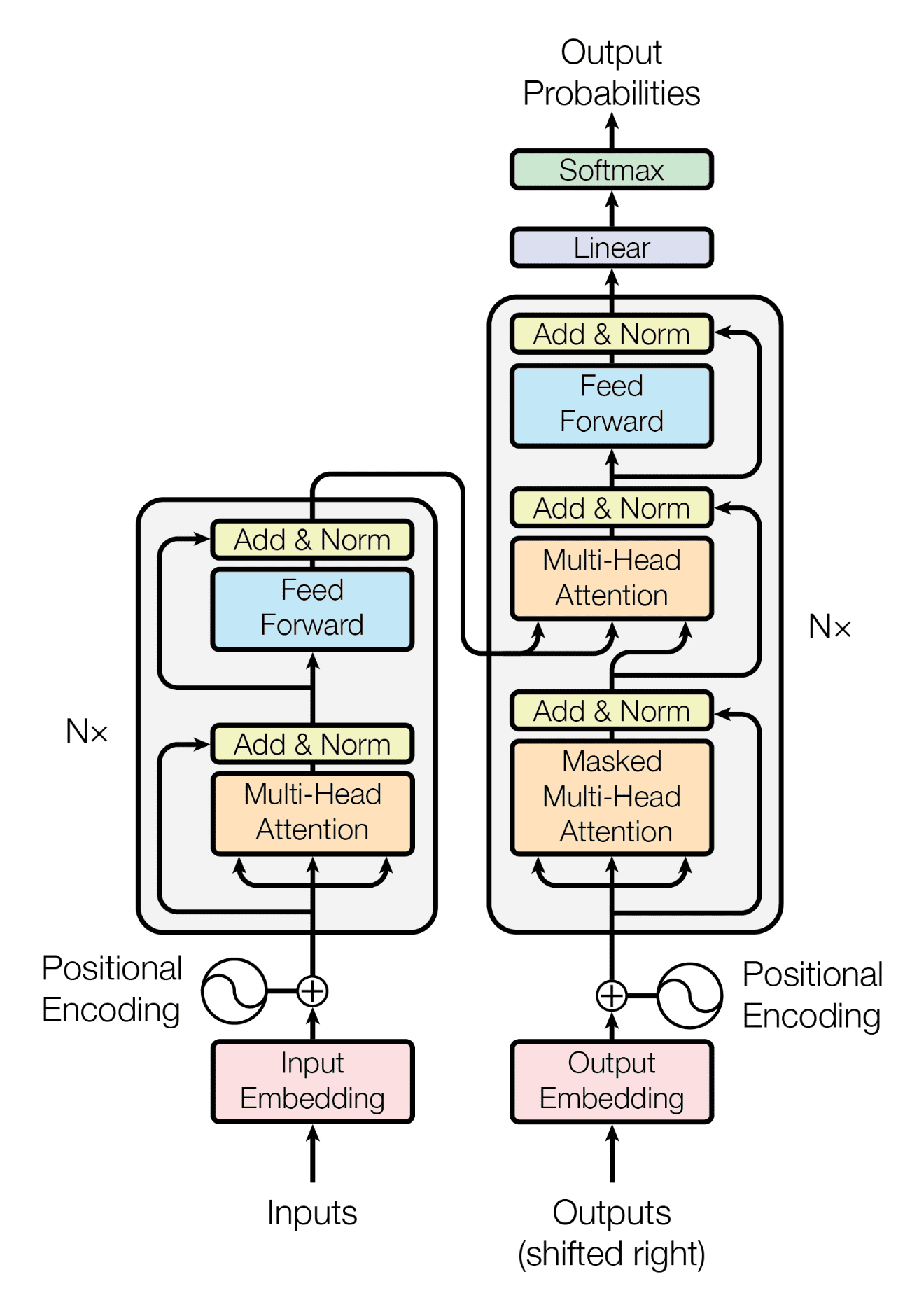}}
\caption{Illustration of a Transformer architecture as presented in \cite{NIPS2017_3f5ee243}}
\label{transformer}
\end{figure}

The LSTM is a sequential model in which one element of the sequence is processed at a time, which is not the case for the Transformer, illustrated in Fig.\ref{transformer}. The Transformer is a Deep Learning model (i.e. a neural network) of the seq2seq type (takes a sequence as input and returns a sequence as output) which has the particularity of only using the attention mechanism and no recurrent or convolutional network. The Transformer is more efficient in tracking remote dependencies but is however more data intensive.

To implement them, we used the OpenNMT library\cite{klein-etal-2017-opennmt} which is an open source ecosystem for neural machine translation and neural sequence learning.
Our implemented LSTM consists of two layers of encoders and two layers of decoders with 500 hidden units, an embedding of size 500 and a dropout layer associated with a rate of 0.3. As optimizer we defined the Stochastic Gradient Descent\cite{sgd} with a learning rate of 1.0 as recommended in the OpenNMT documentation \footnote{\url{https://opennmt.net/OpenNMT-py/options/train.html}}. We added a global attention layer \cite{luong2015effective} which is a simplification of the "classic" attention mechanism proposed in \cite{bahdanau2016neural} and which may achieve better results.
Regarding the Transformer-based architecture, we have reproduced the same features as those defined in the base paper (Vanilla Transformer) \cite{NIPS2017_3f5ee243}. To adapt it to our low-resource context, we reduced some parameter values such as the number of training steps, the early-stopping threshold, the number of validation steps and the warmup steps.

We then divided our dataset into train, validation and test sets with 140,000 sentences for the train set and 7,000 sentences for each of the remaining sets. We applied stratified sampling to ensure that each subset is representative of the overall dataset. This enables a fairer evaluation of the model's performance and limits the biases that can creep into the process. We then applied two different types of segmentation\footnote{Task of dividing a text into coherent and semantically meaningful segments} on the data: SentencePiece\cite{kudo-richardson-2018-sentencepiece} and Character-Level Subwording which has been shown in \cite{lee-etal-2017-fully} to be very effective in translation tasks. All models are trained until convergence, which we consider as reached when no improvement on the validation set is observed after 04 epochs.  In the case where the model does not converge, we set a limit of 30k epochs to stop the training. We then compared the performance of the models on the raw data (not subworded) and on the subworded data to evaluate the impact of this process on the models performance. The vocabularies used on the datasets are generated on all segments of training sets and models are evaluated with Accuracy\footnote{The percentage of correct predictions made by a model} as a metric calculated on test sets at a sentence level. All experiments took place on a virtual machine with a Tesla V-100 GPU with 16GB of RAM.

\section{Results and Perspectives}\label{results}
We compared the two models on the same dataset in different subwording scenarios and Table \ref{accuracy-tab} shows their performances in accuracy given in percentage. We notice that the LSTM model greatly outperforms the Transformer one when no subwording is applied with accuracies of 50.09\% and 9.46\% respectively which are the lowest performances. We observe a similar pattern with the SentencePiece subworded data where the LSTM still outperforms the Transformer with an accuracy of 69.14\% versus 6.99\%. The highest scores were achieved with character-level subworded data where the Transformer performed the best with an accuracy of 81\% compared to the LSTM which achieved an accuracy of 77.67\%. In fact, when no tokenization is applied, it tends to reduce the size of the vocabulary to the total number of words in the corpus. This has the effect of limiting the occurrence of words and reduces the ability of the model to learn these words\cite{tokenization}. The models are thus very sensitive to rare or out-of-vocabulary words (OOV), which results in the generation of $\texttt{<unk>}$ tags during predictions and greatly hinders models' capabilities. In addition, since the Transformer has significantly more parameters than the LSTM, it requires much more data to capture the most error patterns. This may explain why it performs poorly than LSTM under these conditions. This problem is therefore alleviated by tokenizing into subwords or at a character level and the latter has the great advantage to make the model usable on all languages. Furthermore, since most spelling errors occur at the character level (omission, addition and replacement), a model that processes text under these conditions will have a greater ability to capture these kinds of errors.

\begin{table}[htbp]
\centering
\caption{Performance of LSTM and Transformer models evaluated with Accuracy on synthetic Wolof data depending on the type of subwording applied.}
\label{accuracy-tab}
\resizebox{\textwidth}{!}{%
\begin{tabular}{|l|ll|}
\hline

\multicolumn{1}{|c|}{\textbf{Model Architecture}} & \multicolumn{2}{c|}{\textbf{Accuracy (\%)}} \rule{0pt}{4ex}\\ \hline
\multirow{3}{*}{LSTM} & \multicolumn{1}{l|}{\textbf{No Subword}} & \textbf{50.09} \rule{0pt}{4ex}\\ \cline{2-3} 
 & \multicolumn{1}{l|}{\textbf{SentencePiece}} & \textbf{69.14} \rule{0pt}{4ex}\\ \cline{2-3} 
 & \multicolumn{1}{l|}{Character-level} & 77.67 \rule{0pt}{4ex}\\ \hline \hline
\multirow{3}{*}{Transformer} & \multicolumn{1}{l|}{No Subword} & 09.46 \rule{0pt}{4ex}\\ \cline{2-3} 
 & \multicolumn{1}{l|}{SentencePiece} & 06.99 \rule{0pt}{4ex}\\ \cline{2-3} 
 & \multicolumn{1}{l|}{\textbf{Character-level}} & \textbf{81.00} \rule{0pt}{4ex}\\ \hline
\end{tabular}%
}
\end{table}

Table \ref{lstm-char-examples} and \ref{tf-char-examples} show some predictions of the LSTM and Transformer models on character-level subworded data. We notice that both models are able to learn the errors while considering the context. This can be attributed to the attention module which integrates information from surrounding words into the embedding of the current word. The other advantage of these models over dictionary-based models is that they are very robust to out-of-vocabulary words. They are also scalable in the sense that their performance increases as they are used when live data is collected back, corrected and then re-injected as training data, which makes them very powerful. However, we note that the LSTM model fails in some cases such as the last two rows of Table \ref{lstm-char-examples} where it seems to have trouble correcting accents while the Transformer model did a perfect job on the considered extract. The latter nevertheless still seems to have concerns about handling accents as shown in Table \ref{tf-char-examples}. The second row of this table also illustrates an error on the reference side, which suggests the presence of artifacts in the training data that may explain this phenomenon. The last two rows illustrate an interesting problem related to the rule engine. Indeed, the latter identifies the patterns to be processed without distinction while exceptions such as proper nouns (e.g. Kuchner, last line of the table) or some common nouns such as Espagnol (second last line) could mislead the model.

\begin{table}[htbp]
\centering
\caption{Qualitative evaluation of the Character-level LSTM predictions on few conventional Wolof inputs along with corresponding correction (prediction) expected outputs (reference).}
\label{lstm-char-examples}
\begin{tabular}{|l|l|l|}
\hline
\multicolumn{1}{|c|}{\textbf{Input}} & \multicolumn{1}{c|}{\textbf{Prediction}} & \multicolumn{1}{c|}{\textbf{Reference}} \rule{0pt}{4ex}\\ \hline
Ndieekhitaloum diouin bi & Njeexitalum juin bi & Njeexitalum juin bi \rule{0pt}{4ex}\\ \hline
Daa nourou kou beg & Daa nuru ku bég & Daa nuru ku bég \rule{0pt}{4ex}\\ \hline
Dougnou leen dakh & Duñu leen dàq & Duñu leen dàq \rule{0pt}{4ex}\\ \hline
Gnoo and ak orob & Ñoo ànd ak \textbf{orob} & Ñoo ànd ak \textbf{órób} \rule{0pt}{4ex}\\ \hline
Bignouy oubbi bank bi & Biñuy ubbi \textbf{bank} bi & Biñuy ubbi \textbf{bànk} bi \rule{0pt}{4ex}\\ \hline
\end{tabular}%
\end{table}

\begin{table}[htbp]
\setlength{\tabcolsep}{10pt}
\renewcommand{\arraystretch}{1}
\caption{Qualitative evaluation of the Character-level Transformer predictions on few official Wolof sentences (model's outputs) along with references (expected outputs)}
\begin{center}
\begin{tabular}{c c}
\hline\rule{0pt}{12pt}
\textbf{Reference} & \textbf{Predictions} \\[2pt]
\hline\rule{0pt}{12pt}
Nitu Lóot ya weddi woon nañu & Nitu \textbf{Loot} ya wéddi woon nañu \\
\hline\rule{0pt}{12pt}
\textbf{Yeen} a nu muccal nun ñépp & Yéen a nu muccal nun ñépp \\
\hline\rule{0pt}{12pt}
Sama robóo multi la tudd & Sama \textbf{roboo} multi la tudd \\
\hline\rule{0pt}{12pt}
Yàkkamti naa dégg liñuy wax & \textbf{Yakkamti} naa dégg liñuy wax \\
\hline\rule{0pt}{12pt}
Mu gestu ci samag wall & Mu gestu ci samag \textbf{wàll} \\
\hline\rule{0pt}{12pt}
\textbf{Espagnol} bi dooleel na ku ñuul ki & \textbf{Español} bi dooleel na ku ñuul ki
 \\
\hline\rule{0pt}{12pt}
Na dem te yóbbaale Bhl ak Kouchner & Na dem te yóbbaale Bhl ak \textbf{Kuchner} \\[2pt]
\hline
\end{tabular}
\label{tf-char-examples}
\end{center}
\end{table}

This paper is an initial work opening the way to investigate Deep Learning based approaches to address the spelling correction problem in Wolof. The Transformer model already shows promising performances and can be further improved to better adapt to low-resource scenarios as studied in \cite{araabi-monz-2020-optimizing}.
In perspective, we will further analyze the nature of the errors made by the model in order to study the appropriate solutions.

We will also improve our noisy data generator by collaborating with linguists to better identify common errors and create corresponding rules. The e-learning platform of the WAX group of linguists would be very useful to collect data from dictated exercises performed by students. A similar approach has been taken in \cite{mizumoto-etal-2011-mining} which has resulted in a high quality, real-world corpus. We will also explore unsupervised approaches to learn common errors from a noisy corpus instead of a rule engine like the one used in this paper. We also plan to extend the polyglot language identification model on the collected synthetic data to improve its performance in detecting conventional Wolof. This will allow us to later scrape real data from social networks, have it corrected by linguists and then use the resulting parallel corpus to fine-tune our spelling correction model on it. This is particularly important in order to take into account sensitive phenomena such as code-switching, which refers to the passage from one language to another in the same conversation. This phenomenon is very characteristic of everyday Wolof which is strongly influenced by French. We will thus explore NLP approaches addressing this phenomenon of code-switching as studied in \cite{cetinoglu-etal-2016-challenges} in order to make the model more robust to real-world cases. We will also explore other tokenization mechanisms specific to Wolof that could be more efficient than the Character-Level one used here and extend the current system to a model that can make the correspondence in both directions between the two forms of writing.

\section{Conclusion}\label{concl}
We presented the first dataset for spelling correction in Wolof to date, as well as the first approach that addresses the issue from a Deep Learning and Machine Translation perspective. The corpus contains 154,000 sentences, making with their French equivalents the largest French/Wolof parallel corpus collected to date. As the collection is still in progress, the datasets are not yet publicly available.
In addition, we have performed experiments on the two most used NLP architectures, namely the LSTM and the Transformer, on the collected synthetic data. We implemented these architectures using the OpenNMT library and built baseline models. We evaluated these models based on the Accuracy metric computed at a sentence level and we compared their performance regarding the type of subwording applied to the data. We then showed that the Vanilla Transformer model used on character-level subworded data performed the best. We ended by proposing possible improvements that could broaden the scope of such systems and greatly boost their performance.

We have also shown that such a system is crucial for the proper working of NLP applications for Wolof that are being built and will be built in the future. It could also be a major asset in the adoption of the written form of Wolof through large-scale integration into the keyboards of smartphones and other devices.

%
%
%
\bibliographystyle{spbasic_unsrt}
\bibliography{mybibliography}

\end{document}